\documentclass[letterpaper, 10 pt, conference]{ieeeconf}  

\IEEEoverridecommandlockouts                              

\usepackage{booktabs}  
\usepackage{graphicx}  
\usepackage{array}     
\usepackage[table]{xcolor}
\usepackage{xcolor}
\usepackage{amsmath} 
\usepackage{float}
\usepackage{amsfonts}
\usepackage{subcaption} 
\usepackage[font={small}]{caption}
\usepackage{booktabs}
\usepackage{multirow}
\usepackage{multicol}
\usepackage{makecell}
\usepackage{calc}
\usepackage{pifont} 
\usepackage{tikz}
\usepackage{caption}

\title{\LARGE \bf
CVD-SfM: A Cross-View Deep Front-end Structure-from-Motion System for Sparse Localization in Multi-Altitude Scenes \vspace{-4mm}

\thanks{{\footnotesize{Y. Li, Y. Huang, B. Gaudel, H. Jafarnejadsani, and B. Englot are with Stevens Institute of Technology, Hoboken, NJ, USA, \{\texttt{yli21}, \texttt{yhuang85}, \texttt{bgaudel}, \texttt{hjafarne}, \texttt{benglot}\}\texttt{@stevens.edu}. 
}}}}

\author{
Yaxuan Li, Yewei Huang, Bijay Gaudel, Hamidreza Jafarnejadsani, and Brendan Englot
}

\begin{document}

\maketitle
\thispagestyle{empty}
\pagestyle{empty}

\begin{abstract}
We present a novel multi-altitude camera pose estimation system, addressing the challenges of robust and accurate localization across varied altitudes when only considering sparse image input. The system effectively handles diverse environmental conditions and viewpoint variations by integrating the cross-view transformer, deep features, and structure-from-motion into a unified framework. To benchmark our method and foster further research, we introduce two newly collected datasets specifically tailored for multi-altitude camera pose estimation; datasets of this nature remain rare in the current literature. The proposed framework has been validated through extensive comparative analyses on these datasets, demonstrating that our system achieves superior performance in both accuracy and robustness for multi-altitude sparse pose estimation tasks compared to existing solutions, making it well suited for real-world robotic applications such as aerial navigation, search and rescue, and automated inspection.
\end{abstract}\vspace{-1mm}

\section{INTRODUCTION}
Structure-from-motion (SfM)~\cite{visualsfm, colmap, moulon2016openmvg} has been receiving extensive attention in the field of computer vision and robotics; it is pivotal in various real-world applications, including autonomous navigation~\cite{9954049}, rapid mapping after natural disasters for situational awareness, detailed preservation of historical landmarks~\cite{his}, and immersive virtual reality (VR) experiences. As research progresses, it provides a cornerstone in achieving pose estimation and reconstruction, standing out as a particularly effective technique, specifically when input is limited to sparse images. 

While conventional SfM approaches deliver impressive results under conditions of abundant image overlap, they often struggle with sparse input captured at vastly different altitudes. In these scenarios, the drastic viewpoint differences limit shared visual features, making it difficult to establish reliable correspondences. For example, limited overlap between drone and ground images often leads to poor or failed reconstruction \cite{pixsfm} \cite{Shi_2023_ICCV}, highlighting the need for a robust
SfM framework addressing large viewpoint variations and
sparse overlaps.

\begin{figure}
    \centering
    \begin{subfigure}[b]{\linewidth}
        \centering
        \includegraphics[width=0.8\linewidth]{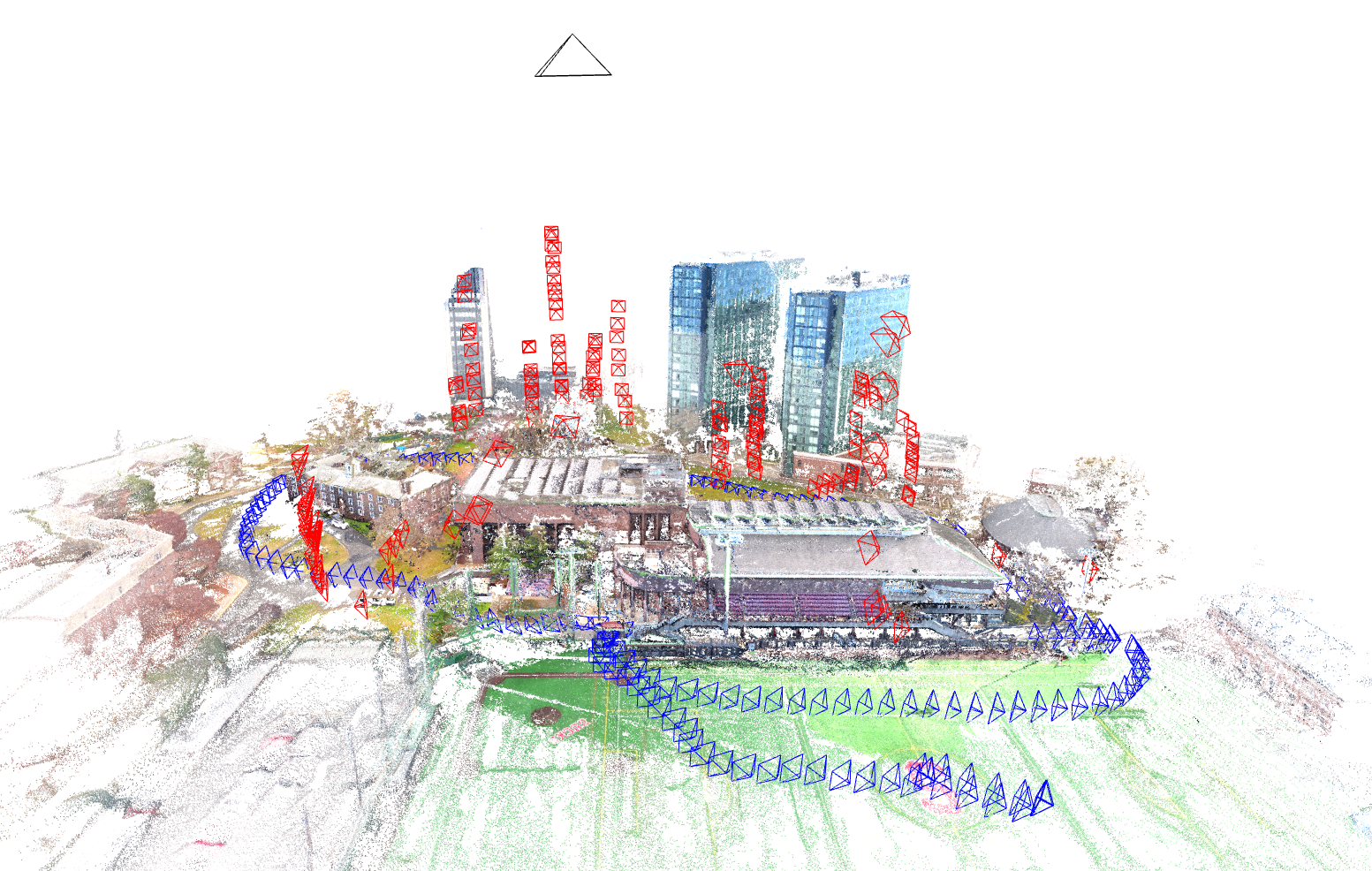}
        \caption{\textbf{Camera poses, and scene reconstruction, from a CVD-SfM solution.} Estimated aerial poses are shown in red, ground poses are shown in blue, and satellite pose (altitude not to scale) is shown in  black.}
        \label{fig:overview_top}
    \end{subfigure}
    
    \vspace{0.3cm} 

    \begin{subfigure}[b]{\linewidth}
        \centering
        \includegraphics[width=0.7\linewidth]{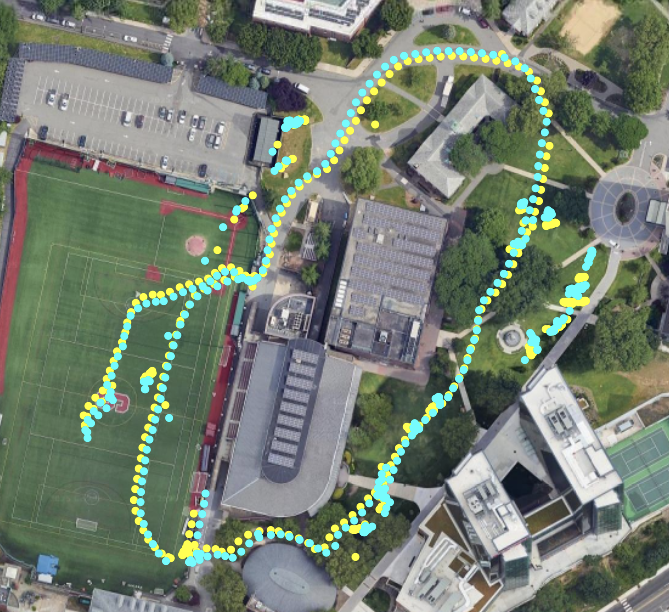}
        \caption{\textbf{Ground truth camera poses} (cyan) alongside estimated poses from CVD-SfM (yellow) from our SIT campus dataset.}
        \label{fig:overview_bottom}
    \end{subfigure}

    \caption{\textbf{CVD-SfM solution over our Stevens Institute of Technology campus dataset.} Pose estimation and scene reconstruction are shown at top, with a ground truth comparison at bottom.}
    \label{fig:overview} \vspace{-7mm}
\end{figure}

Recent research efforts have begun to address the challenge of large viewpoint variations and reduced image overlap. However, existing approaches focus primarily on the front-end of SfM, attempting to enhance feature extraction and matching to handle complex scenarios \cite{Jin_2020} \cite{hloc}. Alternatively, some approaches accelerate correspondence generation and improve accuracy by employing image retrieval and related techniques to obtain similar image pairs before feature extraction. When handling multi-altitude tasks, these methods can sometimes improve the resulting accuracy. However, experiments have shown that due to their strong matching capabilities, they may also introduce incorrect correspondences, which can degrade the final results. Moreover, merely improving the front-end may help find more correspondences when the viewpoint changes, but in cases with minimal overlap, its impact is almost negligible. 


In this paper, we introduce CVD-SfM, an enhanced and robust multi-altitude pose estimation system. 
CVD-SfM (Cross-View Deep feature Structure-from-Motion) applies a cross-view transformer before the traditional SfM pipeline,
enabling the propagation of the geometric information from satellite images into the pose estimation process.
The resulting 3-DoF position estimates for ground and aerial images also provide an initial guess for reconstruction and geometric constraints in the Bundle Adjustment process.
Additionally, we compare learning-based and hand-crafted front-ends within our pipeline using multi-altitude datasets and select DISK+LightGlue as the optimal choice. \textcolor{black}{Fig.~\ref{fig:overview} shows a representative result over one of our datasets.}

Our contributions are summarized as follows:  
\begin{itemize}  
    \item We present the first framework, to the best of our knowledge, to incorporate information from satellite images to support geo-referenced 6-DoF multi-altitude camera pose estimation and comprehensive 3D reconstruction.
    \item Our proposed algorithm surpasses state-of-the-art methods in multi-altitude pose estimation tasks.
    \item We introduce two new publicly available multi-altitude datasets, comprising images from three altitude levels — ground, aerial, and satellite — along with corresponding GPS ground truth information.
\end{itemize}  

The subsequent sections of this paper are organized as follows. 
Sec.~\ref{sec:related}~reviews recent research trends in pose estimation and scene reconstruction.
Sec.~\ref{sec:method} gives a detailed description of our proposed CVD-SfM algorithm.
The experimental results are presented in Sec.~\ref{sec:experiment}, and Sec.~\ref{sec:conclusion} concludes the paper with a summary. 
{\color{black} Our code and data are avaliable at \textbf{https://github.com/RobustFieldAutonomyLab/CVD-SfM}}. 

\vspace{-2mm}

\section{RELATED WORK}\label{sec:related}
The major challenge that makes the multi-altitude pose estimation problem difficult is the limited number of feature correspondences due to viewpoint change.
One key problem is how to find the correct feature correspondences when given images with a large viewpoint change and a wide baseline. Traditional SfM pipelines rely on hand-crafted front-ends, utilizing SIFT (Scale-Invariant Feature Transform)~\cite{article} to extract and describe keypoints, 
and identifying matches based on the ratio test of mutual nearest neighbors. 
However, hand-crafted front-ends always struggle to incorporate enough of the correct correspondences when dealing with images from different altitudes \cite{Sarlin_2020_CVPR}.

Recently, breakthroughs in deep learning have expanded the possibilities for feature extraction and matching. 
Researchers have explored deep learning-based front-ends to enhance SfM performance.
Jin et al.~\cite{Jin2020ImageMA} present a local feature detection and matching benchmark for images across wide baselines, analyzing their impact on SfM and camera pose estimation. 
Their study finds that while deep learning-based methods yield promising results, they require extensive hyperparameter tuning and are more sensitive to dataset variations.
However, this benchmark relies on photo-tourism datasets and does not account for large-scale environments and extreme viewpoint variations. 

GTSfM~\cite{baid2023distributed} is a modular, parallelizable, and distributed global SfM system that integrates deep learning-based feature matching techniques into global SfM. 
However, it struggles in environments with highly repetitive structures (e.g., buildings with similar facades).
Furthermore, although 
learning-based two-view correspondence estimation increases point density in global SfM reconstruction,  SIFT-based methods like COLMAP~\cite{colmap} still achieve higher accuracy. 
HF-Net~\cite{hloc} is a hierarchical localization approach that efficiently estimates the 6-DoF pose of a camera within a known 3D model using a coarse-to-fine approach: first retrieving candidate locations (global image retrieval), then refining poses through local feature matching. 
Although HF-Net achieves high accuracy and speed, it relies on a pre-built sparse SfM model for localization.
In addition, it trades accuracy for efficiency by using hierarchical retrieval instead of direct 2D-3D feature matching. 

Another key problem is lacking correspondences due to the minimal overlap between images captured from different altitudes.
When captured from significantly different viewpoints, images may have minimal or no overlap, making correspondence estimation challenging.
For example, a drone capturing a bird's-eye view of a building primarily sees its rooftop, whereas a ground vehicle observes the building's facade, resulting in little to no overlapping visual information.

Many recent approaches focus on cross-view geolocalization, which estimates the location of a ground image in a corresponding satellite image. 
Zhang et al.~\cite{10031017} introduce an end-to-end sequence-based cross-view geolocalization pipeline. Instead of using single images, their method processes sequences of ground images with limited field of view (FOV) and matches them to corresponding aerial images.
However, this method fails with sparse image input and is restricted to scenes with roads.
PureACL~\cite{10378602} is a fine-grained visual-only cross-view localization method for 
dynamic environments using sparse
matching between ground and satellite images.
However, this method requires initial pose estimations and is sensitive to camera configurations. 
HC-Net~\cite{wang2024fine} is also a fine-grained cross-view geolocalization approach.
While HC-Net achieves meter-level GPS accuracy
on the VIGOR and KITTI datasets, it relies on known camera intrinsics and may struggle in featureless scenes. 
Shi et al.~\cite{Shi_2023_ICCV} present a geometry-guided cross-view transformer that estimates the rotation and translation parts separately.
However, similar to other cross-view geolocalization approaches, 
this method is limited to 3-DoF pose estimation, estimating only the horizontal position and yaw. 

In addition, there are some other SfM pipelines worth our notice. VGGSfM~\cite{vggsfm} leverages an end-to-end differentiable pipeline with deep tracking, global camera estimation, and learnable triangulation, making it more robust in wide-baseline, low-texture, and large viewpoint variation scenarios. However, it struggles with large-scale multi-altitude datasets due to scalability constraints, as it lacks the incremental expansion capability of traditional SfM methods. The end-to-end global optimization requires significant computational resources, and its deep tracking model, optimized for unordered images, might face difficulties handling extreme viewpoint differences caused by altitude variations. Liu et al.~\cite{hybrid} proposed a strategy to enhance incremental SfM by integrating points, lines, and vanishing points (VP) as hybrid features, leading to more robust and accurate reconstructions, especially in low-texture and structured environments. However, in multi-altitude datasets, drastic viewpoint changes can weaken line and VP constraints, reduce line-matching stability, and introduce challenges in hybrid feature-based pose estimation. 

Inspired by these methods, we propose CVD-SfM, a multi-altitude pose estimation framework that integrates a cross-view transformer and a learning-based front-end into SfM.
We initialize the SfM reconstruction using geometric information from satellite images and incorporate geometric constraints into Bundle Adjustment (BA).
To our knowledge, this is the first framework to achieve accurate, geo-referenced 6-DoF multi-altitude camera pose estimation and comprehensive 3D reconstruction with the aid of satellite imagery.\vspace{-1mm}



\section{CVD-SfM Algorithm}\label{sec:method} 
\begin{figure*}[htbp]
    \centering
    \includegraphics[width=0.8\textwidth]{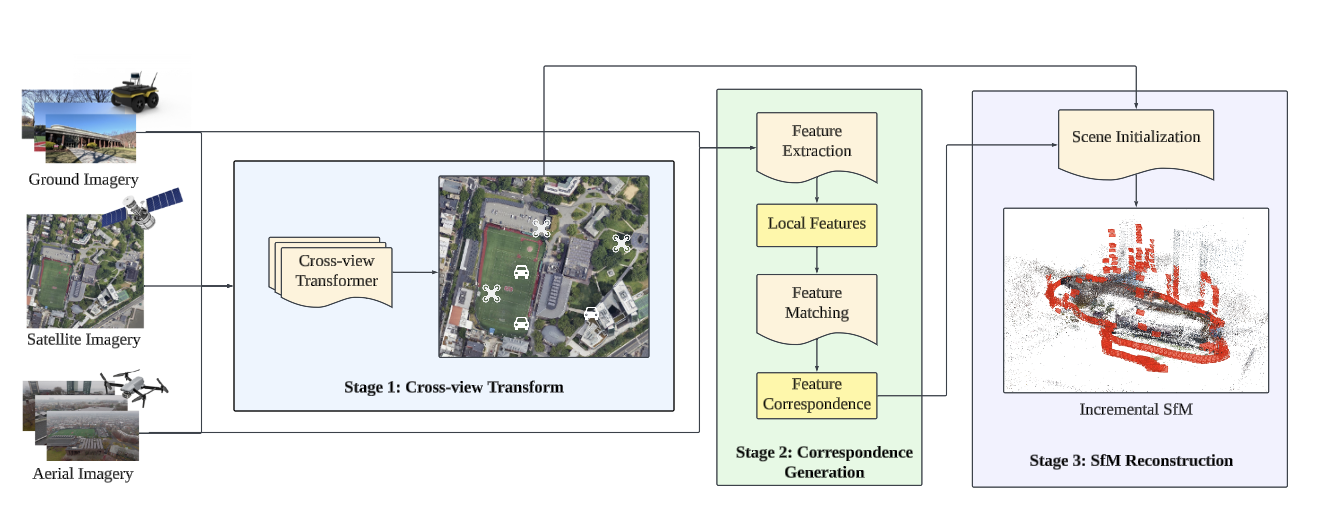}
    \caption{\textbf{Overview of the Proposed CVD-SfM Algorithm}. Each image is transformed into the global frame using satellite images as a geometric reference via the \textbf{cross-view transformer}. The estimated 3-DoF camera poses of aerial and ground images are represented by white drone and vehicle marks, respectively. In the \textbf{correspondence generation} step, DISK \cite{NEURIPS2020_a42a596f} features are extracted and matched using LightGlue \cite{Lindenberger_2023_ICCV}. During \textbf{SfM reconstruction}, both the geometric priors and feature correspondences from the previous steps are used for initialization. Finally, a classical incremental SfM pipeline is applied for pose estimation and reconstruction.}
    \label{fig_method} \vspace{-6mm}
\end{figure*}

Fig.~\ref{fig_method} presents the complete pipeline of the proposed multi-altitude pose estimation system. Given a dataset of images captured at varying altitudes - ground, aerial, and satellite - our system performs pose estimation and reconstruction through three key steps: (1) \textbf{cross-view transformation}, (2) \textbf{correspondence generation}, and (3) \textbf{incremental SfM}.
In the cross-view transformation step, the 2D geometric priors of aerial and ground images are estimated.
Next, the correspondence generation step extracts and associates features across images.
Finally, both the geometric priors and feature correspondences are passed into the incremental SfM for pose estimation and scene reconstruction.

\vspace{-1mm}
\subsection{Cross-View Transformation}\label{sec:cross}
We utilize a geometry-guided cross-view transformer model specifically designed for ground-to-satellite image registration \cite{Shi_2023_ICCV}. 
The cross-view transformation estimates the rotation and translation components of the horizontal 3-DoF camera pose separately.

Rotation estimation begins by synthesizing an overhead-view feature map from ground-view images, where the ground image features are projected into an overhead-view feature map using ground plane homography.
To enhance feature representation, a multi-head self-attention (MHSA) layer contextualizes the overhead-view feature map, followed by a multi-head cross-attention (MHCA) layer that incorporates additional features from the ground-view images for refinement. 
The refined overhead-view features are then passed to a neural optimizer, which estimates the rotation part of the 3-DoF camera pose by comparing them with the actual satellite features. 

For the translation estimation, the overhead feature map is re-synthesized using the estimated rotation and an initial translation set to zero.
This feature map then serves as a sliding window for dense spatial correlation with the satellite features.
Additionally, an uncertainty map representing satellite semantics is encoded to help exclude impossible camera positions.



Finally, the horizontal geometric information, $\mathcal{T}_i = [x_i^{\mathcal{T}}, y_i^{\mathcal{T}}, \theta_{i}^\mathcal{T}]^{\top}$, is obtained for each image in the dataset. These geometric references serve as priors for initializing and constraints for bundle adjustment in incremental SfM, providing a reliable starting point for optimizing camera poses and reconstructing the scene structure.

\vspace{-1mm}
\subsection{Correspondence Generation}\label{sec:corr}
There are various state-of-the-art feature extraction and matching strategies, among which we select four representative methods for consideration and comparison:
\begin{itemize}
\item \textbf{SIFT+NN}: Extracts SIFT~\cite{article} features and applies the mutual nearest-neighbor (MNN) ratio as the matching criterion.
\item \textbf{DISK+LightGlue}: Utilizes DISK~\cite{NEURIPS2020_a42a596f} features and matches them using 
LightGlue~\cite{Lindenberger_2023_ICCV}, a transformer-based feature matcher.
\item \textbf{Aliked+LightGlue} Extracts Aliked~\cite{aliked} features and matches them using LightGlue.
\item \textbf{SuperPoint+SuperGlue}: A deep feature-based method that employs SuperPoint for feature extraction and SuperGlue, a graph neural network (GNN)-based matcher.
\end{itemize}

Our selection is based on recent benchmarks on deep features. Zhao et al.~\cite{aliked} and Dusmanu et al.~\cite{multiview} performed comprehensive benchmarking of state-of-the-art deep features, and their studies demonstrated that SuperPoint, DISK and Aliked were superior over others, especially in viewpoint changing scenarios. Thus, we choose them as representative deep feature extractors. For a fair comparison, all models are trained by the original authors. The performance of these approaches is evaluated on both public and self-collected datasets within the full proposed framework in Sec.~\ref{sec:experiment}.

\vspace{-1mm}
\subsection{Incremental Structure-from-Motion (SfM)}
\textcolor{black}{Incremental SfM deserves more attention than batch methods in robotics due to its potential to handle real-time, sequential data processing, making it more suitable for online applications, robust to incremental updates, and offering better adaptability to dynamic environments. In this paper, we propose a COLMAP-style incremental SfM pipeline for relative pose estimation and reconstruction.}
\subsubsection{Initialization}

The estimated geometric information of all images $\mathcal{T} = \{\mathcal{T}_i\}$ from Sec.~\ref{sec:cross} and the feature correspondences $\mathcal{C}$ computed in Sec.~\ref{sec:corr} are used to initialize the scene for reconstruction. 
Each horizontal camera pose $\mathcal{T}_i \in \mathcal{T}$ serves as prior information for the corresponding 6-DoF pose.
We select the initial image pair \( (I^*, J^*) \) based on three key criteria: 

\noindent\textbf{Maximizing Feature Correspondences}. The image pair with the highest number of feature correspondences is selected for initialization:
\vspace{-1mm}
\begin{equation}
(I^*_i, J^*_i) = \arg\max_{(I_i, I_j)} match(I_i, I_j),
\vspace{-2mm}
\end{equation} 
with the number of valid correspondences between images \( I_i \) and \( I_j \) defined by:
\vspace{-1mm}
\begin{equation}
match(I_i, I_j) = \sum_{k} 1(\text{match}(\mathbf{x}_i^k, \mathbf{x}_j^k)), 
\vspace{-2mm}
\end{equation}  
where \( \mathbf{x}_i^k \) and \( \mathbf{x}_j^k \) are corresponding keypoints in \( I_i \) and \( I_j \), and \( 1(\cdot) \) is an indicator function that counts valid feature matches.  

\noindent\textbf{Epipolar Constraint}.
All selected image pairs must be validated using the epipolar geometry constraint:
\vspace{-1mm}
\begin{equation}
\mathbf{x}_j^T E \mathbf{x}_i = 0.
\vspace{-2mm}
\end{equation}
\( E \) is the essential matrix;
\( \mathbf{x}_i\) and \( \mathbf{x}_j \) are the matched feature points in homogeneous coordinates.
A robust estimation of \( E \) is computed using RANSAC by minimizing the residuals of the epipolar constraint:
\vspace{-1mm}
\begin{equation}
E^* = \arg\min_{E} \sum_k \rho\left( \mathbf{x}_j^T E \mathbf{x}_i \right).
\vspace{-2mm}
\end{equation}
Here, \( \rho(\cdot) \) is the Tukey loss function~\cite{turkey}, which reduces the influence of outliers.

\subsubsection{Image Registration}
After initialization, the reconstruction is expanded using a next-best-view (NBV) strategy.  
To enhance robustness, an efficient uncertainty-driven NBV selection is proposed.
The next-best-view is the candidate image $I_k$ that optimally balances feature visibility and pose uncertainty:
\vspace{-2mm}
\begin{equation}
    I_k^* = \arg\max_{I_k} \left( \sum_{j} w_{j} \cdot \mathbb{I} ( \mathbf{x}_{kj} \in I_k ) - \lambda \cdot \text{Unc}(I_k) \right),
    \vspace{-1mm}
\end{equation}
where $\mathbf{x}_{kj}$ denotes the 2D observation of the 3D point $\mathbf{X}_j$ in image $I_k$, and $w_j$ represents the confidence weight of each feature. The indicator function $\mathbb{I}(\cdot)$ returns 1 if the feature is observed and 0 otherwise. $\lambda$ is a regularization parameter, and $\text{Unc}(I_k)$ is the pose uncertainty calculated from the Perspective-n-Point (PnP) covariance.

\subsubsection{Sparse Reconstruction}
To account for potential false positives in the geometric priors estimated by the cross-view transformation, these priors are jointly optimized with visual features within the SfM framework.
The influence of geometric prior constraints is dynamically adjusted based on the confidence level of visual feature matches.


\noindent\textbf{Pose Estimation with Geometric Priors}.
To estimate the relative pose between the new camera pose $C_{new}$ and the set of known camera poses $C_{known}$, 
both intrinsics and feature correspondences are needed.
To estimate the initial relative pose, the essential matrix is first derived from the fundamental matrix using the intrinsic matrix. 
The singular value decomposition (SVD) of the essential matrix is then used to recover the relative rotation and translation.
\textcolor{black}{To ensure applicability of our system, we do not require EXIF information or known intrinsic parameters. An approximate focal length is assumed for initialization:}
\vspace{-1mm}
\begin{equation}
    f_{\text{init}} = 1.2 \cdot \max(H, W),
    \vspace{-1mm}
\end{equation}
where $H$ and $W$ are the image height and width, respectively.

With the geometric prior $\mathcal{T}_i$ obtained from the cross-view transformer, the initial camera pose is set as:
\begin{align}
{R}_i^{\text{init}} &= R_{\psi}(\theta_{i}^\mathcal{T})\\
{t}_i^{\text{init}} &= \begin{bmatrix} x_i^{\mathcal{T}} & y_i^{\mathcal{T}} & z_i \end{bmatrix}^{\top}.
\end{align}
Here, \( R_{\psi}(\theta_{i}^{\mathcal{T}}) \) denotes the rotation matrix corresponding to the yaw angle \( \theta_{i}^{\mathcal{T}} \).
The height component \( z_i \) remains unconstrained.
The initial poses of all images are therefore defined in the global frame.


\noindent\textbf{Multi-View Triangulation}.
Unlike two-view triangulation, which can be solved analytically in some cases, multi-view triangulation is formulated as a nonlinear least-squares optimization problem:
\vspace{-1mm}
\begin{equation}
    \mathbf{X}^* = \arg\min_{\mathbf{X}} {\sum_{i} \left\| \begin{bmatrix} u_i \\ v_i \end{bmatrix} - \frac{1}{w_i}
    \begin{bmatrix} x_i \\ y_i \end{bmatrix} \right\|^2},
    \vspace{-1mm}
\end{equation}
where $\mathbf{X} = [X, Y, Z, 1]^\top$ is the homogeneous 3D point, $\mathbf{x}_i = [u_i, v_i, 1]^\top$ is the observed 2D projection, and 
\begin{equation}
\begin{bmatrix} x_i \\ y_i \\ w_i \end{bmatrix} = \mathbf{P}_i \mathbf{X},
\end{equation}
with $\mathbf{P}_i$ denoting the projection matrix and $w_i$ denoting the depth component after projection.
The optimization is performed using the Levenberg-Marquardt (LM) algorithm to iteratively refine the 3D point estimate.

\noindent\textbf{Geometry-Aware Bundle Adjustment}. 
In this work, we integrate geometric prior constraints from the cross-view transformation and jointly optimize intrinsic parameters within the Bundle Adjustment (BA) framework.
Our BA optimizes camera poses in the global frame, intrinsic parameters, and 3D structure by minimizing a nonlinear least-squares objective using the LM algorithm:
\vspace{-1mm}
\begin{align}
\mathbf{T}_i^{*}, \mathbf{K}_i^{*}, \mathbf{X}_i^{*} = \arg\min_{\mathbf{T}_i, \mathbf{K}_i, \mathbf{X}_i}  \mathcal{L}_{r} + \mathcal{L}_{\mathcal{T}}.
\vspace{-2mm}
\end{align}
The cost function $\mathcal{L}$ consists of two terms: the reprojection loss $\mathcal{L}_{r}$ and the geometric consistency constraint $\mathcal{L}_{\mathcal{T}}$. \textcolor{black}{$\mathcal{L}_{r}$ ensures visual consistency between observed 2D points and the projected 3D points, and $\mathcal{L}_{\mathcal{T}}$ penalizes deviations from the cross-view geometric priors.}

The reprojection loss $\mathcal{L}_{r}$ considers both intrinsic $\mathbf{K}_i$ and extrinsic $\mathbf{T}_i$ parameters: 
\vspace{-1mm}
\begin{equation}
\mathcal{L}_{r} = \sum_{i} \sum_{j} \| \pi(\mathbf{K}_i, \mathbf{T}_i, \mathbf{X}_j) - \mathcal{C}_{ij} \|^2,
\vspace{-2mm}
\end{equation}
where $\pi(\cdot)$ is the projection of 3D point $X_j$ into the image plane of camera $i$, and $\mathcal{C}_{ij}$ is the observed 2D feature correspondence in image $j$.

To mitigate the influence of false positives from the cross-view transformer, the geometric consistency constraint is introduced:
\vspace{-1mm}
\begin{align}
\mathcal{L}_{\mathcal{T}} &= 
\lambda_t \sum_{i} w_i \cdot \| (X_i, Y_i)^\top - (x_i^{\mathcal{T}}, y_i^{\mathcal{T}})^\top \|^2  \nonumber \\
&+ \lambda_R \sum_{(i,j) \in \mathcal{E}} w_{ij} \cdot \| (\mathbf{R}_{ij} - \mathbf{R}_{ij}^{\mathcal{T}}) \|_F^2\nonumber\\
&+\lambda_{m} \sum_{(i,j) \in \mathcal{E}} \| (\mathbf{T}_i - \mathbf{T}_j) - (\mathbf{T}_i^{\mathcal{T}} - \mathbf{T}_j^{\mathcal{T}}) \|^2  \\
w_i^t &= \frac{1}{1 + \alpha N_{\text{matches}, i}}, 
w_{ij}^R = \frac{1}{1 + \beta N_{\text{matches}, ij}}.
\end{align} 

$\mathcal{L}_{\mathcal{T}}$ constraint consists of three terms: the translation constraint, the rotation constraint, and the relative motion constraint, with corresponding scaling factors $\lambda_t$, $\lambda_R$, and $\lambda_m$. $(X_i, Y_i)^\top$ represents the horizontal position of the camera pose estimated by SfM, while $(x_i^{\mathcal{T}}, y_i^{\mathcal{T}})^\top$ serves as the geometric prior from the cross-view transformation. $\mathbf{R}{ij}$ and $\mathbf{R}{ij}^\mathcal{T}$ denote the relative rotation matrices from SfM and the cross-view transformation, respectively, with their difference measured using the Frobenius norm $\|\mathbf{R}{ij} - \mathbf{R}{ij}^\mathcal{T}\|_F$. $\mathbf{T}^{\mathcal{T}}_i$ represents the full 6-DoF transformation from the cross-view prior.

These terms are combined with confidence-aware weights $w_i^t$ and $w_{ij}^R$, which adjust the influence of the geometric prior based on the number of strong feature matches, $N_{\text{matches}, i}$ and $N_{\text{matches}, ij}$.
\textcolor{black}{Therefore, the influence of prior knowledge is automatically reduced when reliable feature correspondences are available, thus avoiding overfitting on noisy priors. In contrast, under sparse correspondences, the geometric priors have more control and preserve the global structure.}
\section{EXPERIMENTS}\label{sec:experiment}
\subsection{Third-Party and Custom-Gathered Datasets}
To evaluate the performance of our pipeline, we need cross-view datasets that contain images from different altitudes, and there should be at least several images at each altitude. Unfortunately, there are not many publicly available datasets like this at the moment. Recently, thanks to the rise of cross-view geolocalization, there are some cross-view datasets containing satellite and street imagery, like Cross-view KITTI and Ford Multi-AV datasets~\cite{Shi_2023_ICCV}, VIGOR~\cite{9578740}, CVUSA~\cite{salem2020dynamic} and CVACT~\cite{8954224}. These datasets contains sequential street views and satellite images of the surrounding area, which have limited altitude diversity due to the lack of low-altitude aerial imagery, and are not ideal for evaluating our pipeline. University-1652~\cite{zheng2020university} provides views from ground, drone, and satellite for 1652 buildings of 72 universities around the world, but it does not provide GPS information. The Zurich Urban Micro Aerial Vehicle Dataset~\cite{doi:10.1177/0278364917702237}, also known as AGZ (Fig.~\ref{fig_agz}), contains street-level imagery and low-altitude aerial imagery. The aerial imagery is captured by a GoPro Hero 4 camera mounted on a micro-aerial vehicle (MAV) flying within urban streets at low altitudes (5-15 meters above the ground). The resolution is $1920 \times 1080$. The ground-level street-view images are from the Google Street View API, with resolution of $640 \times 360$. The dataset also includes accurate GPS ground truth, which is ideal to evaluate and benchmark our pipeline.

Accordingly, to better evaluate our performance and benchmark state-of-the-art pose estimation pipelines, we introduce two cross-view datasets which contain ground imagery, low-altitude aerial imagery, and satellite imagery. Each dataset focuses on buildings with complex structure. One is a joint building of the Samuel C. Williams Library and the Schaefer Athletic and Recreation Center of Stevens Institute of Technology (Fig.~\ref{fig1}), and the other is a gazebo structure located at Raritan Bay Waterfront Park in South Amboy, NJ (Fig.~\ref{fig2}). The ground imagery was collected by a GoPro 13 with resolution $3840 \times 2160$, and drone imagery was collected by GoPro 11 with resolution $2656 \times 1494$,  mounted on a Skydio X10 drone equipped with a GPS receiver. For all ground-level imagery, we successfully applied RTK GNSS using two Emlid Reach RS+ receivers to obtain accurate ground-level GPS information. The datasets contain the following quantities of ground and aerial images:
\begin{figure}[t]
\centering
\includegraphics[width=\linewidth]{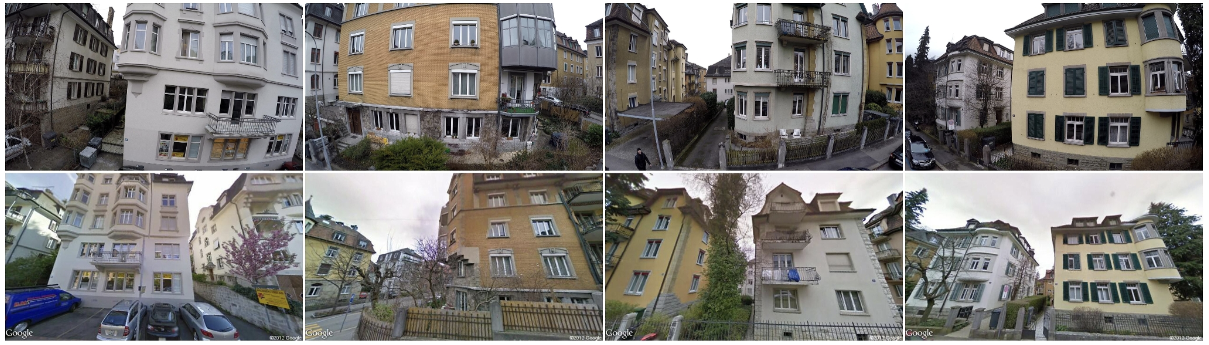}
\caption{\textbf{AGZ Dataset Overview.} Top: low-altitude aerial imagery; bottom: ground imagery.}
\label{fig_agz}\vspace{-4mm}
\end{figure}



\begin{figure}[t]
    \centering
    \begin{subfigure}[b]{0.49\linewidth}
        \centering
        \includegraphics[width=\linewidth]{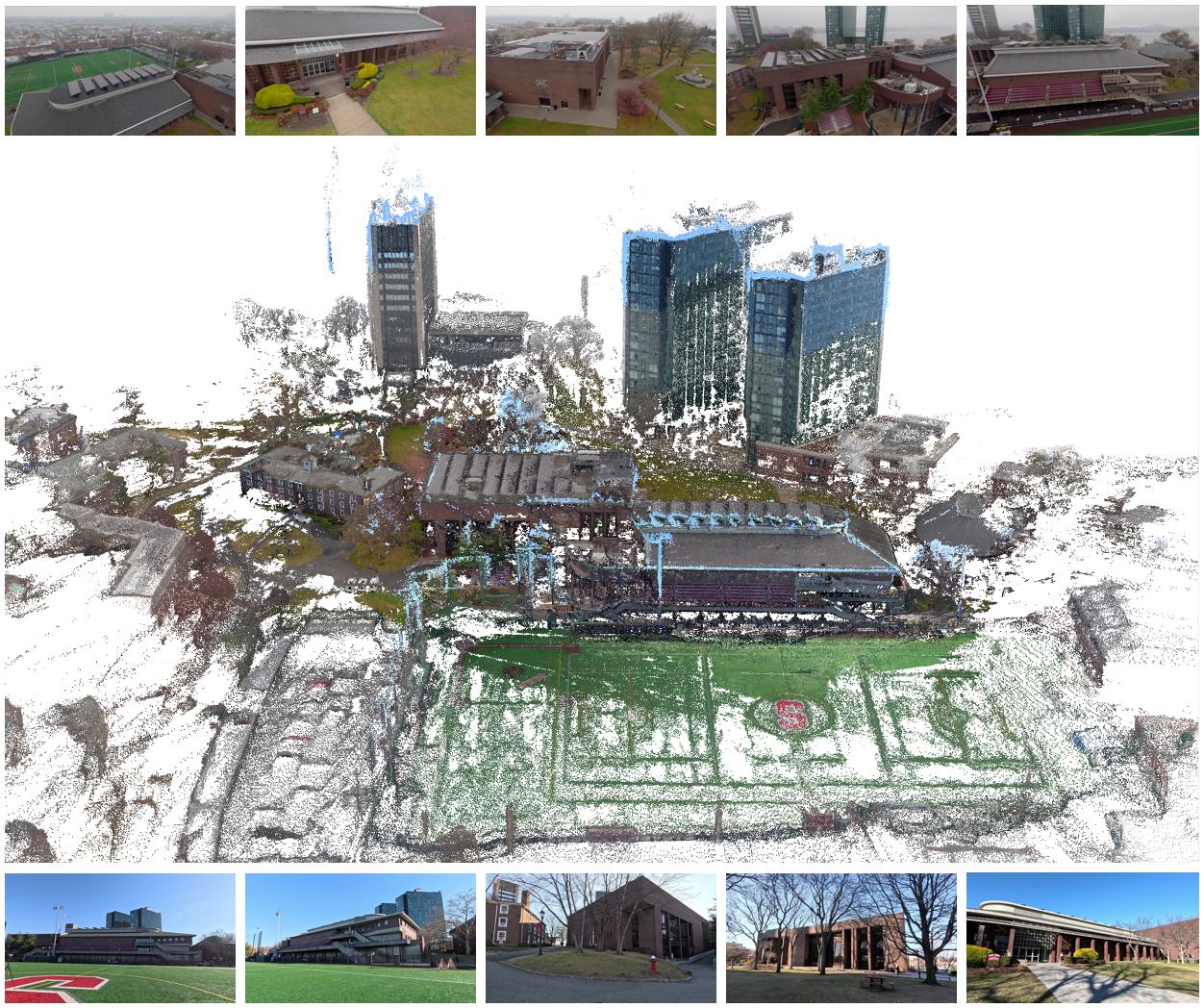}
        \caption{SIT Campus Dataset}
        \label{fig1}
    \end{subfigure}
    \hfill
    \begin{subfigure}[b]{0.49\linewidth}
        \centering
        \includegraphics[width=\linewidth]{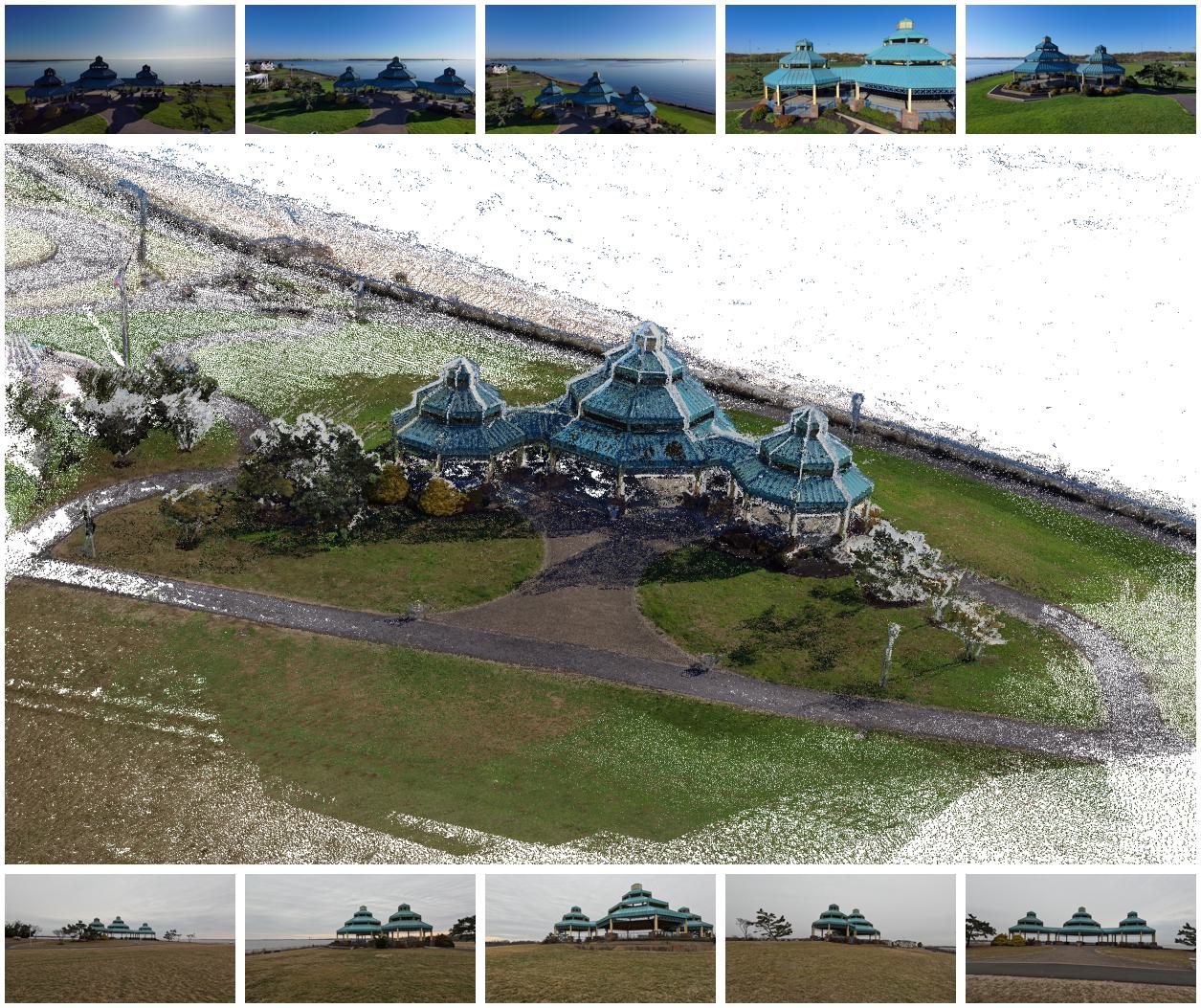}
        \caption{Raritan Bay Park Dataset}
        \label{fig2}
    \end{subfigure}
    \caption{\textbf{Overview of our two custom-gathered datasets:} (a) Stevens Institute of Technology and (b) Raritan Bay Waterfront Park. 
    Example aerial imagery is shown at top, and example ground imagery is shown at bottom.}
    \label{fig:datasets}
    \vspace{-6mm}
\end{figure}

\begin{itemize}
    \item \textbf{SIT Campus:} 186 ground and 179 aerial images.
    \item \textbf{Raritan Bay Park:} 139 ground and 174 aerial images.
    \item \textbf{AGZ:} 50 ground and 157 aerial images.
\end{itemize}
One satellite image for each site was obtained using Google Earth \textcolor{black}{Studio, with resolution of 1920 $\times$ 1080}.

\begin{table*}[htbp]
    \caption{\textbf{Pose Accuracy and Coverage for CVD-SfM, its variants, COLMAP and OpenMVG across three datasets.} }
    \centering
    \renewcommand{\arraystretch}{1.2}
    \resizebox{\textwidth}{!}{
        \begin{tabular}{>{\centering\arraybackslash}p{3cm} | ccc | ccc | ccc | ccc | ccc | ccc }
            \toprule
            \rowcolor{gray!15} \textbf{Framework} & \multicolumn{6}{c|}{\textbf{SIT Campus}} & \multicolumn{6}{c|}{\textbf{Raritan Bay Park}} & \multicolumn{6}{c}{\textbf{AGZ}} \\
            \rowcolor{gray!15} & \multicolumn{3}{c|}{RMSE (m)} & \multicolumn{3}{c|}{Coverage} & \multicolumn{3}{c|}{RMSE (m)} & \multicolumn{3}{c|}{Coverage} & \multicolumn{3}{c|}{RMSE (m)} & \multicolumn{3}{c}{Coverage} \\
            \rowcolor{gray!15} & Min & Mean & Max & Aerial & Ground & Total & Min & Mean & Max & Aerial & Ground & Total & Min & Mean & Max & Aerial & Ground & Total \\
            \midrule
            CVD-SfM  & 0.395 & 4.185 & 36.058 & 99.44\% & 100\% & 99.73\% & 1.803 & 18.464 & 47.725 & 100\% & 100\% & 100\% &  0.291 & 3.481 & 20.123 & 100\% & 98\% & 99.52\% \\
            \rowcolor{gray!15} CVD-SfM Variant A  & 0.511 & 4.07 & 7.356 & 89.94\% & 100\% & 95.07\% & 5.366 & 25.692 & 51.247 & 100\% & 100\% & 100\% & 6.385 & 25.432 & 59.17 & 100\% & 98\% & 99.52\% \\
            CVD-SfM Variant B & 2.237 & 22.521 & 55.162 & 98.32\% & 100\% & 99.18\% & 0.898 & 24.845 & 55.239 & 100\% & 100\% & 100\% & 5.710 & 27.339 & 91.700 & 100\% & 100\% & 100\% \\ 
            \rowcolor{gray!15} CVD-SfM Variant C  & 2.703 & 13.47 & 258.829 & 87.71\% & 97.85\% & 92.88\% & 3.463 & 28.349 & 72.325 & 100\% & 100\% & 100\% & 2.603 & 37.367 & 75.52 & 100\% & 100\% & 100\% \\
            COLMAP  & 0.137 & 1.287 & 4.283 & 58.66\% & 0\% & 28.76\% & 2.002 & 4.332 & 8.26 & 17.24\% & 0\% & 9.58\% & 3.482 & 7.601 & 14.648 & 5.10\% & 0\% & 3.86\% \\
            \rowcolor{gray!15} OpenMVG  &  2.189  & 12.689 & 41.379  & 0\% & 55.91\% & 28.49\% & 0.832 & 4.95 & 9.727 & 13.22\% & 0\% & 7.35\% & 2.906 & 3.77 & 4.708 & 2.55\% & 0\% & 1.93\% \\
            \bottomrule
        \end{tabular}
    }
    \label{tab:benchmark} \vspace{-3mm}
\end{table*} 
\vspace{-2mm}
\subsection{Implementation Details}
We conduct all benchmarks on a desktop workstation equipped two NVIDIA Titan RTX GPUs with 24GB of VRAM and an Intel Core i9-9900K processor with 8 cores and 128GB of RAM, running Ubuntu 20.04. For all experiments, images will be the only input and metadata like intrinsic parameters will not be provided. \textcolor{black}{Additionally, all models used in this study were trained by their original authors - we have not performed any additional training.} \textcolor{black}{To ensure a fair comparison, extensive parameter tuning efforts were carried out for COLMAP and OpenMVG.}
\vspace{-1mm}
\subsection{Evaluation Metrics}
Our evaluation metrics contain two parts: pose accuracy and reconstruction quality. To assess the accuracy and completeness of the estimated camera poses, we evaluate the results using both quantitative errors and coverage metrics. Specifically, we consider positional error and pose coverage (number of successfully estimated poses). For reconstruction quality, since ground truth 3D points are unavailable, we adopt the reprojection error, a common metric in Structure-from-Motion (SfM) and camera calibration.

\noindent\textbf{Pose Positional Error.}
Given a set of estimated camera positions \( \mathbf{X} = \{x_i\} \) and corresponding ground truth GPS positions \( \mathbf{Y} = \{y_i\} \), we apply the Kabsch-Umeyama algorithm to compute the optimal similarity transformation that aligns the estimated pose coordinates with the ground truth:
\vspace{-3mm}
\begin{equation}
    \mathbf{Y} = s \mathbf{R} \mathbf{X} + \mathbf{t},
    \vspace{-2mm}
\end{equation}
where \( \mathbf{R} \in \mathbb{R}^{3 \times 3} \) is the rotation matrix, \( s \in \mathbb{R} \) is the uniform scaling factor, \( \mathbf{t} \in \mathbb{R}^{3} \) is the translation vector.

After aligning the estimated poses with the world coordinate frame, we compute the positional errors using the Root Mean Square Error (RMSE):
\vspace{-1mm}
\begin{equation}
    \text{RMSE} = \sqrt{\frac{1}{N_{\text{estimated}}} \sum_{i=1}^{N} \|\mathbf{y}_i - \hat{\mathbf{y}}_i\|^2},
    \vspace{-1mm}
\end{equation}
where $\mathbf{N_{\text{estimated}}}$ is the number of \textcolor{black}{successfully reconstructed} camera poses, $\mathbf{y}_i$ represents the ground truth camera position, and $\hat{\mathbf{y}}_i$ represents the transformed estimated position.

\noindent\textbf{Pose Coverage.}
We define coverage as the ratio of successfully estimated poses to the total number of input poses:
\vspace{-2mm}
\begin{equation}
\text{Coverage Rate} = \frac{N_{\text{estimated}}}{N_{\text{input}}},
\vspace{-1mm}
\end{equation}
where $N_{\text{input}}$ is the total number of camera poses, and $\hat{\mathbf{y}}_i$ represents the transformed estimated position.
A high coverage rate indicates robust pose estimation performance, while a low coverage rate suggests that many camera poses could not be reliably estimated. \textcolor{black}{A camera pose is considered successfully reconstructed if it is estimated and included in the final model, regardless of the positional error magnitude.}

\noindent\textbf{Reprojection Error.}
Given a set of $N$ image feature points $\mathbf{p}_i = (u_i, v_i)$ detected in an image and their corresponding estimated 3D points $\mathbf{P}_i = (X_i, Y_i, Z_i)$ in the reconstructed scene, we define the reprojection error as:
\vspace{-2mm}
\begin{equation}
    e_{\text{reproj}} = \frac{1}{N} \sum_{i=1}^{N} \left\| \mathbf{p}_i - \hat{\mathbf{p}}_i \right\|^2,
    \vspace{-1mm}
\end{equation}
where $\mathbf{p}_i$ is the observed 2D image point, $\hat{\mathbf{p}}_i$ is the reprojected 2D point from the estimated 3D structure, and $N$ is the total number of feature correspondences.

A lower reprojection error indicates a more geometrically consistent reconstruction, as it suggests that the estimated 3D structure and camera poses correctly align with the observed 2D image points.
\vspace{-1mm}
\subsection{Benchmarks}\label{sec:benchmark}
Our benchmarking includes the classic SfM pipelines, COLMAP~\cite{colmap} and OpenMVG~\cite{moulon2016openmvg}. Additionally, \textcolor{black}{to assess each component’s impact, we conduct ablation studies, consistent with Sec.~\ref{sec:corr}}:
\begin{itemize}
    \item \textbf{CVD-SfM.} Our proposed framework, combining cross-view transformer with Disk and LightGlue.
    \item \textbf{CVD-SfM Variant A  (DISK \& LightGlue).} Excludes cross-view transformer from the pipeline.
    \item \textbf{CVD-SfM Variant B (Aliked \& LightGlue).} Replaces DISK with Aliked on CVD-SfM Variant A. 
    \item \textbf{CVD-SfM Variant C (SuperPoint \& SuperGlue).} Replaces DISK and LightGlue with  SuperPoint and SuperGlue on CVD-SfM Variant A, similar to Pixel-Perfect SfM~\cite{pixsfm}.
    \item \textbf{COLMAP.} Classic {\color{black} incremental} COLMAP, employing SIFT with a match ratio of 0.8.
    \item \textbf{OpenMVG.} {\color{black} Incremental pipeline of OpenMVG}, employing SIFT with a match ratio of 0.8. 
\end{itemize}

\subsection{Results}

\begin{figure*}[htbp]
\centering
\includegraphics[width=0.95\textwidth]{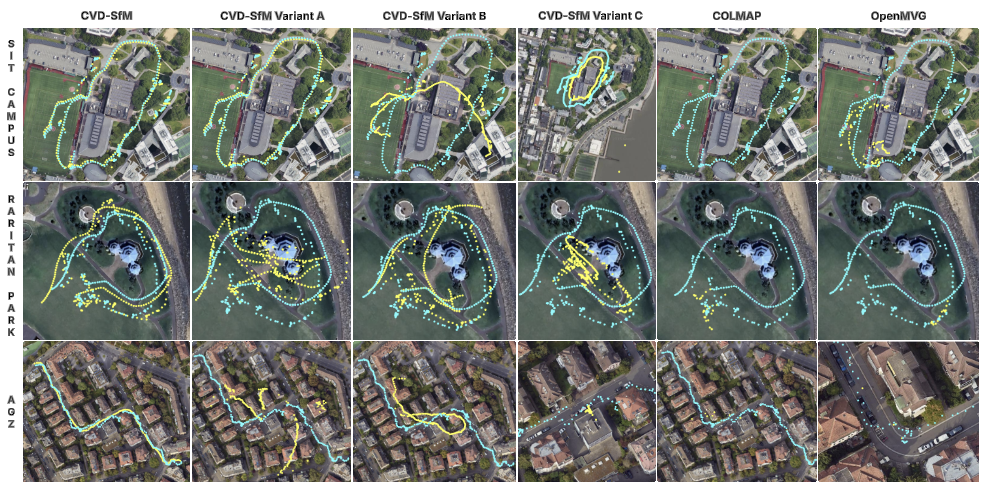}
\captionsetup{justification=centerlast, singlelinecheck=false}
\caption{\textbf{Pose Estimation Visualization.} Yellow Points represent estimated poses; cyan points represent ground truth poses.}
\label{fig_vis} \vspace{-5mm}
\end{figure*}

\textcolor{black}{The average runtime of CVD-SfM across all three datasets is 37 minutes. Variants A and B have similar runtimes, ranging from approximately 25 to 34 minutes, depending on the dataset. Variant C has the longest runtime due to SuperGlue, which involves computationally intensive feature matching and optimization processes. COLMAP and OpenMVG achieve the fastest computation, with runtimes of less than 10 minutes. Fig.~\ref{fig_vis} shows visualizations of pose estimation for all benchmarks across all three datasets.}

\textcolor{black}{For a quantitative comparison,} Table~\ref{tab:benchmark} shows positional error (RMSE) and coverage of different frameworks. \textcolor{black}{We note that RMSE is computed under optimal alignment conditions rather than as a direct Euclidean distance error.} On the three mixed aerial-ground multi-altitude datasets, our proposed CVD-SfM framework with cross-view transformer demonstrates both high reconstruction coverage and low positional errors. It maintains near or fully complete reconstructions while reducing average positional errors, particularly evident in the Raritan Bay Park and AGZ datasets. (Although its maximum error can be relatively large in the SIT Campus dataset, the mean remains low, and coverage is still excellent.) \textcolor{black}{The highest RMSE, 36.058m, occurred in the SIT Campus dataset due to limited visual overlap and large viewpoint differences between ground and aerial views. The CVD-SfM variants without cross-view transformers underscore the importance of utilizing cross-view attention for effective matching. Though not always reducing maximum error, they improve average accuracy and coverage.}

In more challenging datasets (Raritan Park, AGZ), significantly higher mean and max errors occur when the transformer module is absent, indicating that the cross-view transformer is a critical module for reducing mismatches and improving pose accuracy in scenarios with large viewpoint differences. Traditional SfM pipelines (COLMAP, OpenMVG) struggle with large viewpoint differences, failing to match enough cross-view features. Their coverage rates are extremely low, especially for ground images, but for the small subset of images they do reconstruct, the accuracy (RMSE) can be quite decent. Deep feature pipelines achieve higher coverage but can exhibit larger error variance. \textcolor{black}{These findings reveal a key trade-off in SfM: increasing coverage often reduces translation accuracy, as broader matching may introduce lower-quality correspondences. This underscores the necessity of balancing coverage with pose accuracy, particularly when addressing real-world robotic applications.}

These results indicate that incorporating cross-view transformers and deep features significantly improves the reliability and precision of multi-altitude SfM, effectively integrating aerial and ground imagery for large-scale reconstruction.

\vspace{-1mm}
\begin{table}[htp]
\caption{Reprojection Error of CVD-SfM, its variants, COLMAP and OpenMVG across three datasets.}
\label{tab:reproejct}
\centering
\begin{tabular*}{\linewidth}{@{\extracolsep{\fill}} cccc} 
\toprule
\textbf{Framework} & \textbf{SIT} & \textbf{RARITAN} & \textbf{AGZ} \\ 
\midrule
CVD-SfM & 1.296 & 1.165 & 1.125 \\ 
CVD-SfM Variant A & 1.201 & 1.278 & 1.183 \\
CVD-SfM Variant B & 1.203 & 1.084 &  1.384 \\ 
CVD-SfM Variant C & 1.249 & 1.198 &  1.141 \\ 
COLMAP & 0.903 & 0.667 & 0.831 \\ 
OpenMVG & 0.663 & 0.618 & 0.568 \\ 
\bottomrule
\end{tabular*} \vspace{-3mm}
\end{table}

Table~\ref{tab:reproejct} compares the reprojection errors (the lower, the better) of six different SfM frameworks. We can observe that traditional SfM pipelines using SIFT features and a mutual nearest neighbor ratio check (e.g., OpenMVG and COLMAP) consistently achieve lower reprojection errors. In particular, OpenMVG exhibits the most accurate results overall, closely followed by COLMAP. Meanwhile, learning-based methods (CVD-SfM and its variants) generally show slightly higher errors, although they sometimes approach the performance of traditional approaches. This is because hand-crafted methods are explicitly designed to handle a wide range of scenes and lighting conditions, while deep learning methods always have some learning bias. They are still maturing and often need further refinement to outperform well-established classical approaches in every scenario. However, we can see that the introduction of the cross-view transformer in CVD-SfM helps to reduce the error in many cases. 

{\color{black}
The results of our ablation studies (described in Sec.~\ref{sec:benchmark}) can be summarized as follows:
\begin{itemize}
    \item \textbf{Variant A (No transformer)}: Excluding a cross-view transformer significantly increases RMSE, highlighting its role in providing geometric priors for drift correction and guided reconstruction in low-overlap areas.
    \item \textbf{Variants B \& C (Feature Replacements)}: Replacing DISK+LightGlue with Aliked or SuperPoint yields mixed results. Aliked (B) performs comparably to DISK, but with higher RMSE, indicating lower robustness to viewpoint changes. SuperPoint+SuperGlue (C) achieves good coverage but with the most widely-varying errors, likely due to overfitting or poor generalization across altitudes.
    \item \textbf{COLMAP \& OpenMVG}: These techniques perform well only with successful rectification, but struggle on multi-altitude scenes due to insufficient matching correspondences.
\end{itemize}
}

\vspace{-2mm}
\section{CONCLUSIONS}\label{sec:conclusion}
In this paper, we propose CVD-SfM, an accurate and robust pose estimation framework designed for multi-altitude scenes using a cross-view transformer and satellite reference imagery. By utilizing priors from the cross-view transformer for scene initialization and bundle adjustment, with the help of a deep front-end, CVD-SfM achieves a significant improvement in pose estimation accuracy and coverage over traditional SfM pipelines and other state-of-the-art frameworks when processing multi-altitude datasets. Additionally, we introduce two new cross-view datasets with ground-truth GPS into the community, enriching the diversity of the datasets for future research. \textcolor{black}{In our future work, we will enforce quality thresholds for feature correspondences and geometric priors to enhance reconstruction performance, aiming to reduce outlier errors despite potential coverage loss. We also plan to enhance the model's robustness against challenging day/night transitions, and other domain shifts between gallery and probe image sets, including lighting changes and aerial-to-indoor view transitions.}

\vspace{-2mm}







\bibliographystyle{IEEEtran}
\bibliography{bib}

\begin{thebibliography}{10}
\providecommand{\url}[1]{#1}
\csname url@samestyle\endcsname
\providecommand{\newblock}{\relax}
\providecommand{\bibinfo}[2]{#2}
\providecommand{\BIBentrySTDinterwordspacing}{\spaceskip=0pt\relax}
\providecommand{\BIBentryALTinterwordstretchfactor}{4}
\providecommand{\BIBentryALTinterwordspacing}{\spaceskip=\fontdimen2\font plus
\BIBentryALTinterwordstretchfactor\fontdimen3\font minus \fontdimen4\font\relax}
\providecommand{\BIBforeignlanguage}[2]{{%
\expandafter\ifx\csname l@#1\endcsname\relax
\typeout{** WARNING: IEEEtran.bst: No hyphenation pattern has been}%
\typeout{** loaded for the language `#1'. Using the pattern for}%
\typeout{** the default language instead.}%
\else
\language=\csname l@#1\endcsname
\fi
#2}}
\providecommand{\BIBdecl}{\relax}
\BIBdecl

\bibitem{visualsfm}
C.~Wu, ``Towards linear-time incremental structure from motion,'' in \emph{International Conference on 3D Vision (3DV)}, 2013, pp. 127--134.

\bibitem{colmap}
J.~L. Schönberger and J.-M. Frahm, ``Structure-from-motion revisited,'' in \emph{IEEE Conference on Computer Vision and Pattern Recognition (CVPR)}, 2016, pp. 4104--4113.

\bibitem{moulon2016openmvg}
P.~Moulon, P.~Monasse, R.~Perrot, and R.~Marlet, ``Open{MVG}: Open multiple view geometry,'' in \emph{International Workshop on Reproducible Research in Pattern Recognition}.\hskip 1em plus 0.5em minus 0.4em\relax Springer, 2016, pp. 60--74.

\bibitem{9954049}
S.~Jiang, Q.~Li, W.~Jiang, and W.~Chen, ``Parallel structure from motion for {UAV} images via weighted connected dominating set,'' \emph{IEEE Transactions on Geoscience and Remote Sensing}, vol.~60, pp. 1--13, 2022.

\bibitem{his}
F.~Brandolini and G.~Patrucco, ``Structure-from-motion ({SFM}) photogrammetry as a non-invasive methodology to digitalize historical documents: A highly flexible and low-cost approach?'' \emph{Heritage}, vol.~2, pp. 2124--2136, 07 2019.

\bibitem{pixsfm}
P.-E. Sarlin, P.~Lindenberger, V.~Larsson, and M.~Pollefeys, ``Pixel-perfect structure-from-motion with featuremetric refinement,'' \emph{IEEE Transactions on Pattern Analysis and Machine Intelligence}, vol.~47, no.~5, pp. 3298--3309, 2023.

\bibitem{Shi_2023_ICCV}
Y.~Shi, F.~Wu, A.~Perincherry, A.~Vora, and H.~Li, ``Boosting 3-{D}o{F} ground-to-satellite camera localization accuracy via geometry-guided cross-view transformer,'' in \emph{Proceedings of the IEEE/CVF International Conference on Computer Vision (ICCV)}, 2023, pp. 21\,516--21\,526.

\bibitem{Jin_2020}
Y.~Jin, D.~Mishkin, A.~Mishchuk, J.~Matas, P.~Fua, K.~M. Yi, and E.~Trulls, ``Image matching across wide baselines: From paper to practice,'' \emph{International Journal of Computer Vision}, vol. 129, no.~2, p. 517–547, Oct. 2020.

\bibitem{hloc}
P.-E. Sarlin, C.~Cadena, R.~Siegwart, and M.~Dymczyk, ``From coarse to fine: Robust hierarchical localization at large scale,'' in \emph{IEEE/CVF Conference on Computer Vision and Pattern Recognition (CVPR)}, 2019, pp. 12\,716--12\,725.

\bibitem{article}
D.~Lowe, ``Distinctive image features from scale-invariant keypoints,'' \emph{International Journal of Computer Vision}, vol.~60, pp. 91--110, 11 2004.

\bibitem{Sarlin_2020_CVPR}
P.-E. Sarlin, D.~DeTone, T.~Malisiewicz, and A.~Rabinovich, ``Superglue: Learning feature matching with graph neural networks,'' in \emph{IEEE/CVF Conference on Computer Vision and Pattern Recognition (CVPR)}, 2020, pp. 4938--4947.

\bibitem{Jin2020ImageMA}
Y.~Jin, D.~Mishkin, A.~Mishchuk, J.~Matas, P.~Fua, K.~M. Yi, and E.~Trulls, ``Image matching across wide baselines: From paper to practice,'' \emph{International Journal of Computer Vision}, vol. 129, pp. 517 -- 547, 2020.

\bibitem{baid2023distributed}
\BIBentryALTinterwordspacing
A.~Baid, J.~Lambert, T.~Driver, A.~Krishnan, H.~Stepanyan, and F.~Dellaert, ``Distributed global structure-from-motion with a deep front-end,'' 2023. [Online]. Available: \url{https://arxiv.org/abs/2311.18801}
\BIBentrySTDinterwordspacing

\bibitem{10031017}
X.~Zhang, W.~Sultani, and S.~Wshah, ``Cross-view image sequence geo-localization,'' in \emph{IEEE/CVF Winter Conference on Applications of Computer Vision (WACV)}, 2023, pp. 2913--2922.

\bibitem{10378602}
S.~Wang, Y.~Zhang, A.~Perincherry, A.~Vora, and H.~Li, ``View consistent purification for accurate cross-view localization,'' in \emph{IEEE/CVF Int. Conf. on Computer Vision (ICCV)}, 2023, pp. 8163--8172.

\bibitem{wang2024fine}
X.~Wang, R.~Xu, Z.~Cui, Z.~Wan, and Y.~Zhang, ``Fine-grained cross-view geo-localization using a correlation-aware homography estimator,'' \emph{Advances in Neural Information Processing Systems (NeurIPS)}, vol.~36, 2024.

\bibitem{vggsfm}
J.~Wang, N.~Karaev, C.~Rupprecht, and D.~Novotny, ``V{GGS}f{M}: Visual geometry grounded deep structure from motion,'' in \emph{IEEE/CVF Conference on Computer Vision and Pattern Recognition (CVPR)}, 2024, pp. 21\,686--21\,697.

\bibitem{hybrid}
S.~Liu, Y.~Gao, T.~Zhang, R.~Pautrat, J.~L. Schönberger, V.~Larsson, and M.~Pollefeys, ``Robust incremental structure-from-motion with hybrid features,'' in \emph{18th European Conference on Computer Vision (ECCV)}, 2024, p. 249–269.

\bibitem{NEURIPS2020_a42a596f}
M.~Tyszkiewicz, P.~Fua, and E.~Trulls, ``D{ISK}: Learning local features with policy gradient,'' in \emph{Advances in Neural Information Processing Systems (NeurIPS)}, vol.~33, 2020.

\bibitem{Lindenberger_2023_ICCV}
P.~Lindenberger, P.-E. Sarlin, and M.~Pollefeys, ``Light{G}lue: Local feature matching at light speed,'' in \emph{Proceedings of the IEEE/CVF International Conference on Computer Vision (ICCV)}, 2023, pp. 17\,627--17\,638.

\bibitem{aliked}
X.~Zhao, X.~Wu, W.~Chen, P.~C.~Y. Chen, Q.~Xu, and Z.~Li, ``A{LIKED}: A lighter keypoint and descriptor extraction network via deformable transformation,'' \emph{IEEE Transactions on Instrumentation and Measurement}, vol.~72, pp. 1--16, 2023.

\bibitem{multiview}
M.~Dusmanu, J.~L. Sch{\"o}nberger, and M.~Pollefeys, ``Multi-view optimization of local feature geometry,'' in \emph{16th European Conference on Computer Vision (ECCV))}, 2020, p. 670–686.

\bibitem{turkey}
K.~L. Clarkson, R.~Wang, and D.~P. Woodruff, ``Dimensionality reduction for {T}ukey regression,'' in \emph{International Conference on Machine Learning (ICML)}, vol.~97, 2019, pp. 1262--1271.

\bibitem{9578740}
S.~Zhu, T.~Yang, and C.~Chen, ``V{IGOR}: Cross-view image geo-localization beyond one-to-one retrieval,'' in \emph{IEEE/CVF Conference on Computer Vision and Pattern Recognition (CVPR)}, 2021, pp. 5316--5325.

\bibitem{salem2020dynamic}
T.~Salem, S.~Workman, and N.~Jacobs, ``Learning a dynamic map of visual appearance,'' in \emph{IEEE Conference on Computer Vision and Pattern Recognition (CVPR)}, 2020, pp. 12\,435--12\,444.

\bibitem{8954224}
L.~Liu and H.~Li, ``{Lending Orientation to Neural Networks for Cross-View Geo-Localization},'' in \emph{IEEE/CVF Conference on Computer Vision and Pattern Recognition (CVPR)}, 2019, pp. 5617--5626.

\bibitem{zheng2020university}
Z.~Zheng, Y.~Wei, and Y.~Yang, ``University-1652: A multi-view multi-source benchmark for drone-based geo-localization,'' \emph{28th ACM International Conference on Multimedia (MM)}, pp. 1395--1403, 2020.

\bibitem{doi:10.1177/0278364917702237}
A.~L. Majdik, C.~Till, and D.~Scaramuzza, ``The {Z}urich urban micro aerial vehicle dataset,'' \emph{The International Journal of Robotics Research}, vol.~36, no.~3, pp. 269--273, 2017.

\end{thebibliography}

\end{document}